\documentclass[twocolumn,american,csquotes]{article}
\usepackage[T1]{fontenc}
\usepackage{geometry}
\usepackage{babel}
\usepackage{units}
\usepackage{amsmath}
\usepackage{amsthm}
\usepackage{amssymb}
\usepackage{cancel}
\usepackage{graphicx}
\usepackage[unicode=true]
 {hyperref}

\makeatletter
\usepackage{abstract} 
\newcommand{\makeabstract}{\@ifundefined{abstractcontent}{}{\begin{abstract}\abstractcontent\end{abstract}}}
\newcommand{\makefrontmatter}{\if@twocolumn{\twocolumn[\maketitle\makeabstract\vskip2\baselineskip]\saythanks}\else{\maketitle\makeabstract}\fi}
\theoremstyle{remark}
\newtheorem*{acknowledgement*}{\protect\acknowledgementname}

\usepackage{dsfont}
\usepackage [latin1]{inputenc}

\def\one{\mathds{1}}

\makeatother

\usepackage[style=numeric-comp,sorting=none]{biblatex}
\providecommand{\acknowledgementname}{Acknowledgement}

\begin{document}
\title{Information Field Theory and Artificial Intelligence}
\author{Torsten En{\ss}lin$^{1,2}$ \\
{\small{}$^{1}$ Max Planck Institute for Astrophysics, Karl-Schwarzschild-Str.
1, 85748 Garching, Germany}\\
{\small{}$^{2}$ Ludwig-Maximilians-Universität München, Geschwister-Scholl-Platz
1 80539 Munich, Germany}}
\newcommand{\abstractcontent}{\emph{Information field theory} (IFT), the information theory for
fields, is a mathematical framework for signal reconstruction and
non-parametric inverse problems. \emph{Artificial intelligence} (AI)
and \emph{machine learning} (ML) aim at generating intelligent systems
including such for perception, cognition, and learning. This overlaps
with IFT, which is designed to address perception, reasoning, and
inference tasks. Here, the relation between concepts and tools in
IFT and those in AI and ML research are discussed. In the context
of IFT, fields denote physical quantities that change continuously
as a function of space (and time) and \emph{information theory} refers
to Bayesian probabilistic logic equipped with the associated entropic
information measures. Reconstructing a signal with IFT is a computational
problem similar to training a \emph{generative neural network} (GNN)
in ML. In this paper, the process of inference in IFT is reformulated
in terms of GNN training. In contrast to classical neural networks,
IFT based GNNs can operate without pre-training thanks to incorporating
expert knowledge into their architecture. Furthermore, the cross-fertilization
of variational inference methods used in IFT and ML are discussed.
These discussions suggests that IFT is well suited to address many
problems in AI and ML research and application.}
\makefrontmatter

\section{Motivation}

Determining the concrete configuration of a field from measurement
data is an ill-posed inverse problem as physical fields have an infinite
number of \emph{degrees of freedom} (DoF), whereas data sets are always
finite in size. Thus, the data provide a finite number of constraints
for only a subset of the infinitely many DoF of a field. In order
to infer a field, the remaining of its DoF need therefore to be constrained
via prior information. Fortunately, physics provides such prior information
on fields. This information might either be precise, like $\nabla\cdot B=0$
in electrodynamics, or more phenomenological, in the sense that a
field shaped by a certain process can often be characterized by its
$n$-point correlation functions. Having knowledge on such correlations
can be sufficient to regularize the otherwise ill-posed field inference
problem from finite and noisy data such that meaningful statements
about the field can be made.

As a formalism for this,\emph{ information field theory} (IFT) was
introduced \cite{2009PhRvD..80j5005E,2019AnP...53100127E} following
and extending earlier lines of work \cites[e.g.][]{1987PhRvL..58..741B}{lemm2003bayesian}.
IFT is information theory for fields. It investigates the space of
possible field configurations and constructs probability densities
over those spaces in order to permit Bayesian field inference from
data. It has been applied successfully to a number of problems in
astrophysics \cite{2012A&A...542A..93O,2015A&A...575A.118O,2015A&A...581A..59J,2015MNRAS.449.4162I,2015A&A...581A.126S,2015JCAP...02..041D,2018arXiv180405591K,2019A&A...627A.134A,2020A&A...633A.150H,2020A&A...639A.138L,2020arXiv200205218A,2020arXiv200811435A,2021arXiv210201709H,2021A&A...655A..64M},
particle physics \cite{2014PhRvD..89d3505D,2016JCAP...04..030H,2021JCAP...04..071W},
and elsewhere \cite{2012PhRvE..85b1134S,2013PhRvE..87a3308E,2018PhRvE..97c3314L,2019Entrp..22...46K,2021AnP...53300486F}.
Here, the relation of IFT with methods and concepts used in \emph{artificial
intelligence} (AI) and \emph{machine learning }(ML) research are outlined,
in particular with \emph{generative neural networks} (GNNs) and in
the usage of variational inference. The presented line of arguments
summarizes a number of recent works \cite{2016arXiv161208406E,2017Entrp..19..402L,2018arXiv181204403K,2019arXiv190111033K,2021Entrp..23..693K,2021Entrp..23..853F}.

The motivation for this work are twofold. On the one hand, understanding
conceptual relations between IFT, ML, and AI techniques allows to
transfer computational methods between these domains and to develop
synergistic approaches. This article will discuss such. On the other
hand, the current success of \emph{deep learning} techniques for neural
networks has let them appear as a synonym for AI in the public perception.
This has consequences for decisions about which kind of technologies
get scientific funding. The point this paper is trying to make is
that if deep learning qualifies as AI in this respect, then this should
also apply to a number of other techniques, including those based
on IFT.

The paper is organized as following. IFT is briefly introduced in
Sec.\ \ref{sec:Information-Field-Theory} in its most modern incarnation
in terms of standardized, generative models. These are shown to be
structurally similar to GNNs in Sec.\ \ref{sec:Artificial-Intelligence}.
The structural similarity of IFT inference and GNN training problems
allows for a common set of variational inference methods, as discussed
in Sec.\ \ref{sec:Variational-Inference}. Sec.\ \ref{sec:Conclusion-and-Outlook}
concludes on the relation of IFT methods and those used in AI and
ML and gives an outlook on future synergies.

\section{Information Field Theory\label{sec:Information-Field-Theory}}

\subsection{Basics}

IFT allows to deduce fields from data in a probabilistic way. In order
to be able to apply probability theory onto the space of field configurations,
a measure in this space is needed. Although no canonical mathematical
measure on function spaces exists, for IFT applications the usage
of Gaussian process measures \cite{edward2006rasmussen}, which are
mathematically well defined \cite{lassas2004can,saksman2009discretization},
is usually fully sufficient. Gaussian processes can also be argued
to be a natural starting point for reasoning on fields with known
finite first and second order moments, as we will discuss now.

To be specific, let $\varphi:\Omega\rightarrow\mathbb{R}$ be a scalar
field over some domain $\Omega\subset\mathbb{R}^{u}$ and our prior
knowledge on $\varphi$ be the first and second moments of the field,
e.g.
\begin{eqnarray}
\left\langle \varphi^{x}\right\rangle _{(\varphi)} & = & \overline{\varphi}^{x}\text{ and }\\
\left\langle (\varphi-\overline{\varphi})^{x}(\varphi-\overline{\varphi})^{y}\right\rangle _{(\varphi)} & = & \Phi^{xy}\text{ for all }x,y\in\Omega,
\end{eqnarray}
with $\varphi^{x}:=\varphi(x)$ denoting a field value and $\left\langle f(\varphi)\right\rangle _{(\varphi)}:=\int\mathcal{D}\varphi\,\mathcal{P}(\varphi)f(\varphi)$
a prior expectation value for some function $f$ of the field. If
only the first and second field moments are given as prior information,
it follows from the maximum entropy principle that the least informative
probability distribution function encoding this information is a Gaussian
with these moments. Thus using this Gaussian 
\begin{eqnarray}
\mathcal{P}(\varphi|I) & \equiv & \mathcal{G}(\varphi-\overline{\varphi},\Phi)\\
 & := & \frac{1}{\sqrt{|2\pi\Phi|}}\exp\left(-\frac{1}{2}(\varphi-\overline{\varphi})^{\dagger}\Phi^{-1}(\varphi-\overline{\varphi})\right)\nonumber 
\end{eqnarray}
as a prior with background information $I=(\left\langle \varphi\right\rangle _{(\varphi)}=\overline{\varphi},\left\langle (\varphi-\overline{\varphi})(\varphi-\overline{\varphi})^{\dagger}\right\rangle _{(\varphi)}=\Phi)$
is a conservative choice, as it makes the least additional assumptions
about the field except for the moments specified in $I$.

In many applications, however, the field of interest, the signal $s$,
is not a Gaussian field, but may be related to such via a non-linear
transformation. For example in astronomical applications of IFT, the
sky brightness field $s$ is the quantity of interest, which is strictly
positive and therefore can not be a Gaussian field. However, the logarithm
of\emph{ }a brightness can be positive and negative and may therefore
be modeled as a Gaussian process. In such a case one could assign
e.g. $s^{x}(\varphi)=s_{0}\exp(\varphi^{x})$ as a model for a diffuse
(spatially correlated) sky emission component, with $s_{0}$ a reference
brightness, chosen such that for example $\left\langle \varphi^{x}\right\rangle _{(\varphi)}=0$
holds.

Having established a field prior, Bayesian reasoning on the field
$\varphi$, and therefore on the signal of interest $s=s(\varphi)$,
based on some data $d$ and its likelihood $\mathcal{P}(d|\varphi,I)$
is possible. The field posterior
\begin{equation}
\mathcal{P}(\varphi|d,I)=\frac{\mathcal{P}(d|\varphi,I)\mathcal{P}(\varphi|I)}{\mathcal{P}(d|I)}
\end{equation}
is defined as well as the prior and permits to answer questions about
the field, like its most probable configuration $\varphi_{\text{MAP}}=\text{argmax}_{\varphi}\mathcal{P}(\varphi|d,I)$
(MAP = maximum a posteriori), its posterior mean $m=\left\langle \varphi\right\rangle _{(\varphi|d,I)}$,
or its posterior uncertainty dispersion $D=\left\langle (\varphi-m)(\varphi-m)^{\dagger}\right\rangle _{(\varphi|d,I)}$.
IFT exploits the formalism of quantum and statistical field theory
to calculate such posterior expectation values \cite{2009PhRvD..80j5005E,2011PhRvD..83j5014E,2010PhRvE..82e1112E,2017Entrp..19..402L,2021Entrp..23.1652W}.
These formal calculations, however, should not be the focus here.
Instead, it should be the formulation of IFT inference problems in
terms of generative models as these can be interpreted as GNNs.

For this purpose, the likelihood is expressed in terms of a measurement
equation
\begin{eqnarray}
d & = & R(\varphi)+n\text{, with}\label{eq:measurement}\\
R(\varphi) & := & \left\langle d\right\rangle _{(d|\varphi)},\label{eq:response}\\
n & := & d-R(\varphi)\text{, and}\\
\mathcal{P}(d|\varphi,I) & \equiv & \mathcal{P}(n=d-R(\varphi)|\varphi),
\end{eqnarray}
which is always possible if the data can be embedded into a vector
space and the data expectation value $\left\langle d\right\rangle _{(d|\varphi)}$
exists. Here and in the following, we omit the background information
$I$ in probabilities. This rewriting of the likelihood in terms of
a mean instrument response $d'=R(\varphi)$ to the field $\varphi$
and a noise process $\mathcal{P}(n|\varphi)$, which summarizes the
fluctuations around that mean $d'$, allows to regard the data as
the result of a noisy generative process that maps field values $\varphi$
and associated noise realizations $n$ onto data $d$ according to
Eq.\ \ref{eq:measurement}.

In case the instrument response and noise processes are provided for
the signal $s$ instead of the Gaussian field $\varphi$ as $R'(s):=\left\langle d\right\rangle _{(d|s)}$
and $\mathcal{P}(n|s)$, their respective pull backs $R(\varphi):=\left\langle d\right\rangle _{(d|s(\varphi))}=R'(s(\varphi))$
and $\mathcal{P}(n|\varphi):=\mathcal{P}(n|s(\varphi))$ provide the
necessary response and noise statistics with respect to the field
$\varphi$.

All this provides a generative model for the signal $s$ and data
$d$ via $\varphi\hookleftarrow\mathcal{G}(\varphi,\Phi)$, $s=s(\varphi)$,
$n\hookleftarrow\mathcal{P}(n|s)$, and $d=R'(s)+n$, which should
now be standardized. The standardization introduces a generic latent
space that permits better comparison to GNNs used in AI and ML and
simplifies the usage of variational inference methods discussed later
on.

\subsection{Prior Standardization}

Standardization of a random variable $\varphi$ refers to finding
a mapping from a standard normal distributed random variable $\xi\hookleftarrow\mathcal{G}(\xi,\one)$
to $\varphi$ that reproduces the statistics of $\mathcal{P}(\varphi)$.
For a Gaussian field $\varphi$, this is just a mapping of the form
\begin{equation}
\varphi(\xi):=\overline{\varphi}+\Phi^{\frac{1}{2}}\xi,
\end{equation}
where $\Phi^{\frac{1}{2}}$ refers to a square root of $\Phi$, which
always exists for a covariance matrix that is positive definite. For
the large class of band diagonal and therefore translational invariant
covariance matrices $\Phi$, which are very relevant for applications
as we argue below, the square root of $\Phi$ can be explicitly constructed.

\subsection{Power Spectra}

In many signal inference problems, no spatial location is singled
out a priori, before the measurement. This means that the field covariance
between two locations only depends on the distance between these positions,
but not on their absolute positions. Thus, $\Phi^{xy}=C_{\varphi}(x-y)$.
As a consequence of the Wiener-Khinchin theorem, such a translational
invariant field covariance becomes diagonal in harmonic space,
\begin{equation}
\widetilde{\Phi}^{kq}=\mathcal{F}_{x}^{k}\Phi^{xy}(\mathcal{F^{\dagger}})_{y}^{q}=(2\pi)^{u}\delta(k-q)P_{\varphi}(k)=\widehat{P_{\varphi}}^{kq}.
\end{equation}
Here and in the following, $\mathcal{F}$ denotes a harmonic transform
(a $u$-dimensional Fourier transform $\mathcal{F}_{x}^{k}=\exp(ik\cdot x)$
in case of an Euclidean space, as we assume in the following), $\dagger$
the adjoint (complex conjugate and transposed of a matrix or vector),
$P_{\varphi}(k):=\mathcal{F}_{x'}^{k}C_{\varphi}^{x'}$ is the so
called power spectrum of $\varphi$, the Einstein convention for repeated
indices is used, as in $\widetilde{\varphi}^{k}:=\mathcal{F}_{x}^{k}\varphi^{x}\equiv\int dx^{u}\exp(ik\cdot x)\,\varphi(x)$,
and $\widehat{\phi}=\text{diag}(\phi)$ denotes a diagonal operator
in the space of the field $\phi$ with the values of $\phi$ on the
diagonal.

Thanks to this diagonal representation of the field covariance in
harmonic space, an explicit standardization of the field is given
via 
\begin{eqnarray}
\xi & \hookleftarrow & \mathcal{G}(\xi,\one),\\
\varphi & = & \overline{\varphi}+\mathcal{F}^{-1}A_{\varphi}\,\xi,\text{ and }\\
A_{\varphi} & = & \widehat{P_{\varphi}^{\nicefrac{1}{2}}},
\end{eqnarray}
where the latter is an amplitude operator that is diagonal in harmonic
space and that imprints the right amplitudes onto the Fourier modes
of $\varphi$. This can be seen via a direct calculation,
\begin{eqnarray}
\left\langle (\varphi(\xi)-\overline{\varphi})(\varphi(\xi)-\overline{\varphi})^{\dagger}\right\rangle _{(\xi)} & = & \mathcal{F}^{-1}A_{\varphi}\,\left\langle \xi\xi\right\rangle _{(\xi)}A_{\varphi}^{\dagger}\mathcal{F}^{-1\dagger}\nonumber \\
 & = & \mathcal{F}^{-1}A_{\varphi}\,\one A_{\varphi}^{\dagger}\mathcal{F}^{-1\dagger}\nonumber \\
 & = & \mathcal{F}^{-1}\widetilde{\Phi}\mathcal{F}^{-1\dagger}=\Phi\nonumber \\
 & = & \left\langle (\varphi-\overline{\varphi})(\varphi-\overline{\varphi})^{\dagger}\right\rangle _{(\varphi)}.
\end{eqnarray}
In case no direction of the space is singled out a priori, the two-point
correlation function and the power spectrum of $\varphi$ become isotropic,
$\Phi^{xy}=C_{\varphi}(|x-y|)$ and $\widetilde{\Phi}^{kq}=(2\pi)^{u}\delta(k-q)P_{\varphi}(|k|)$,
respectively. In this case only a one dimensional power spectrum needs
to be known. Such power spectra are often smooth functions on a double
logarithmic scale in Fourier space, since any sharp feature in them
would correspond to a (quasi-) periodic pattern in position space,
which would be very unnatural for most signals. Thus, introducing
the logarithmic Fourier space scale variable $\kappa(k):=\ln k/k_{0}$
with respect to\ some reference scale $k_{0}$, we expect
\begin{equation}
\psi(\kappa):=\ln\left(P_{\varphi}(k_{0}e^{\kappa})/P_{0}\right)
\end{equation}
 to be a field itself, in the sense that it is sufficiently smooth.
Here, $P_{0}$ is a pivot scale for the power spectrum.

\subsection{Amplitude Model}

Often, the power spectrum as parameterized through $\psi$ is not
known a priori for a field $\varphi$, but statistical homogeneity,
isotropy, and the absence of long range quasi-periodic signal variations
make a Gaussian field prior for $\psi$ plausible, $\mathcal{P}(\psi)=\mathcal{G}(\psi-\overline{\psi},\Psi)$.
This log-log-power spectrum may exhibit fluctuations $\chi:=\psi-\overline{\psi}$
around a non-zero mean $\overline{\psi}(\kappa)$. The latter might
e.g. encode a preference for falling spectra and therefore for a spatially
smooth field $\varphi$. In this case just another layer for $\chi$
of a standardized generative model has to be added,
\begin{eqnarray}
\eta & \hookleftarrow & \mathcal{G}(\eta,\one)\\
\chi(\eta) & := & A_{\psi}\eta\text{, with}\\
A_{\psi}A_{\psi}^{\dagger} & := & \Psi\text{, and}\\
\psi(\eta) & := & \overline{\psi}+\chi(\eta).
\end{eqnarray}
Again, a prior for a field, here the only one dimensional $\chi(\kappa)$,
is needed. A detailed description of how this \emph{amplitude model
}can be provided efficiently is given by \cite{2020arXiv200205218A}.
This reference also provides a generative model for the case that
the signal domain $\Omega$ is a product of sub-spaces, like position
space and an energy spectrum coordinate, each requiring a different
correlation structure, and the total correlation being a direct product
of those. Assuming a direct product for the correlation structures
might be possible for many field inference problems \cites[as, e.g. in][]{2018A&A...619A.119P}{2020arXiv200205218A}.

\subsection{Dynamical Systems}

Let us take a brief detour to fields shaped by dynamical systems.
Dynamical systems, typically exhibit correlation structures that are
not direct products of the spatial and temporal sub-spaces, as was
proposed above. Here, the full spatial and temporal Fourier power
spectrum $P_{\varphi}(k,\omega)$, with $\omega$ being the temporal
frequency, encodes the full dynamics of a linear, homogeneous, and
autonomous system. For example, a stochastic wave field $\varphi(x,t)$
may follow the dynamical equation
\begin{equation}
\left(\frac{\partial^{2}}{\partial t^{2}}+\eta\frac{\partial}{\partial t}-c^{2}\frac{\partial^{2}}{\partial x^{2}}\right)\varphi(x,t)=\xi(x,t),\label{eq:wave}
\end{equation}
where $c$ is the wave velocity and $\eta$ a damping constant. The
field dynamics are determined by a response operator (or Green's function)
$G$ that is a convolution of the exciting noise field $\xi$ with
a kernel $g$,
\begin{equation}
\varphi=G\,\xi=g*\xi,
\end{equation}
where $*$ denotes convolution. In Fourier-space, this kernel can
be applied by a direct point wise multiplication, $(\mathcal{F}\varphi)^{(k,\omega)}=(\mathcal{F}g)^{(k,\omega)}\,(\mathcal{F}\xi)^{(k,\omega)}$and
is given by
\begin{eqnarray}
(\mathcal{F}g)^{(k,\omega)} & = & (\omega{{}^2}-i\eta\omega-c^{2}k^{2})^{-1}=:P_{G}(k,\omega).
\end{eqnarray}

If the excitation of field fluctuations is caused by a white, stochastic
noise field $\xi\hookleftarrow\mathcal{G}(\xi,\one)$, the resulting
field has a power spectrum of 
\begin{equation}
P_{\varphi}(k,\omega)=|P_{G}(k,\omega)|^{2}P_{\xi}(k,\omega)=\frac{1}{(\omega{{}^2}-c^{2}k^{2})^{2}+\eta^{2}\omega^{2}}.
\end{equation}
In this case, the spectrum is an analytical function in $\omega$
and $k$. This results from Eq.\ \eqref{eq:wave} being a linear,
homogeneous, and autonomous partial differential equation.

Linear integro-differential equations, however, can still be solved
by convolutions, in which case the kernel might not have an analytically
closed form any more, if the equation is still homogeneous and autonomous.
For example in \emph{neural field theory} \cite{nunez1974brain,amari1975homogeneous,amari1977dynamics,coombes2014tutorial},
a macroscopic description of the brain cortex dynamics, the neural
activity $\varphi(x,t)$ might be described by
\begin{equation}
\frac{\partial}{\partial t}\varphi=-\varphi+w*(f\circ\varphi)+\xi.
\end{equation}
Here, $w$ is a spatial-temporal convolution kernel (that usually
contains a delta function in time), $f:\mathbb{R}\rightarrow\mathbb{R}$
an activation function that is applied point wise to the field, $(f\circ\varphi)(x,t)=f(\varphi(x,t))$,
and we added an input term $\xi$. In case $f$ is linear, the system
responds linearly to inputs. Then the input response is a convolution
with a kernel $g$ that has in general a non-analytical spectrum,
\begin{equation}
(\mathcal{F}g)^{(k,\omega)}=\left(1+i\omega-(\mathcal{F}w)^{(k,\omega)}\,f'\right)^{-1},
\end{equation}
where $f'$ is the slope of $f$ and $\mathcal{F}w$ the Fourier transformed
kernel of the dynamics.

An inference of such non-analytical and highly structured response
spectra from data is possible with IFT and can be used to learn the
system dynamics from noisy system measurements \cite{2017PhRvE..96e2104F,2021AnP...53300486F}.
It just requires a more complex spectral prior than discussed here.
Let us now return to our main line of argumentation.

\begin{figure}[t]
\includegraphics[viewport=82bp 60bp 752bp 750bp,clip,width=1\columnwidth]{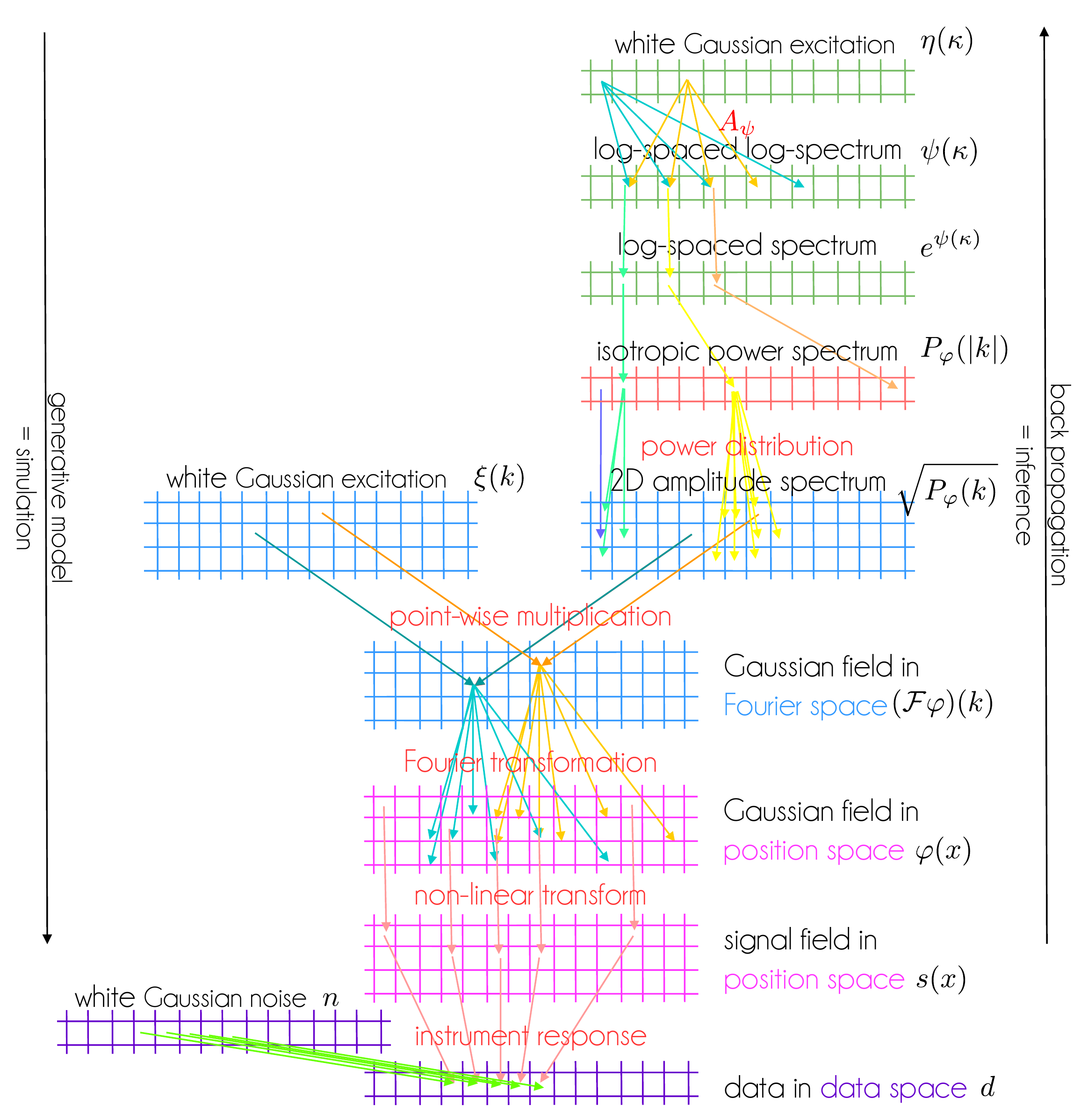}\caption{An IFT model for a 2D Gaussian random field with also generated homogeneous
and isotropic correlation structure and its measurement according
to Eqs.\ \eqref{eq:zeta-generation}-\eqref{eq:data-generation}
displayed as a GNN. Layers with identical shapes are given identical
colors. Note that all layers have a physical interpretation and the
architecture of this GNN encodes expert knowledge on the field. Inserting
random numbers into the latent spaces and executing the network from
top to bottom corresponds to a simulation of signal and data generation.
``Learning'' the latent space variables from bottom to top via back
propagation of data space residuals with respect to to observed data
corresponds to inference. \label{fig:An-IFT-model}}
\end{figure}

\subsection{Generative Model}

\begin{figure*}[t]
\includegraphics[width=1\textwidth]{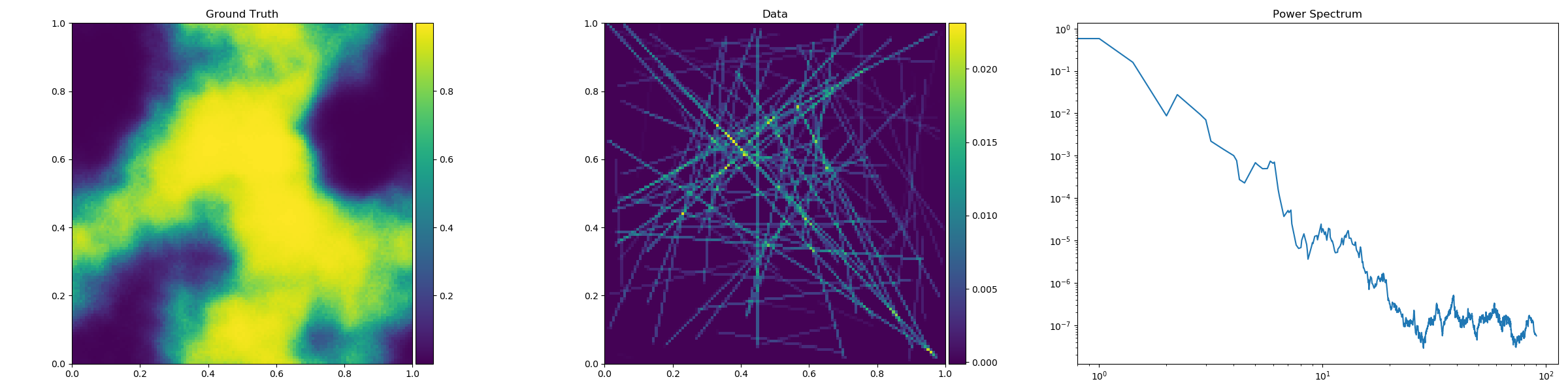}\\
\includegraphics[width=1\textwidth]{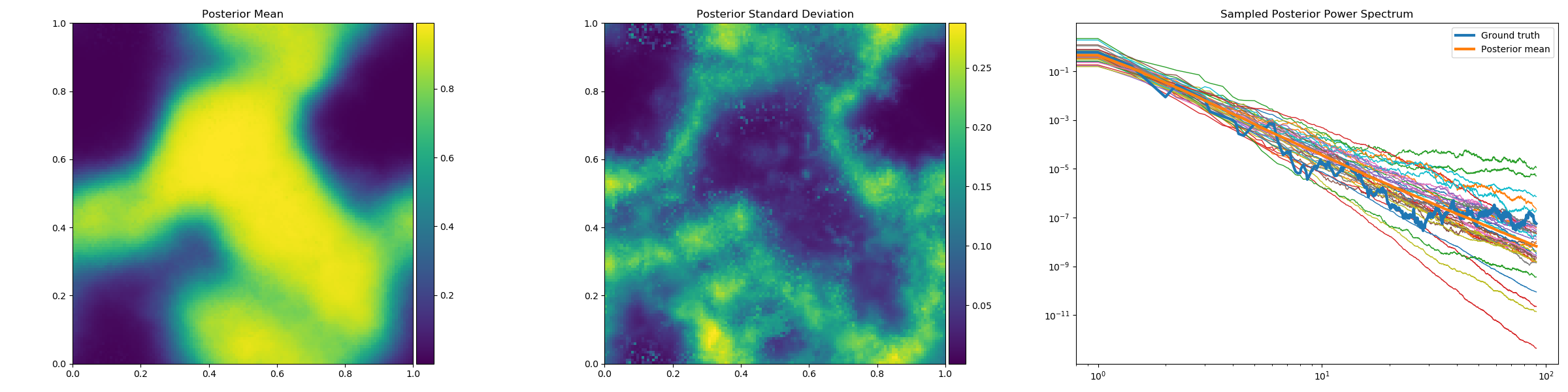}\caption{Output of a generative IFT model for a 2D tomography problem in simulation
(top row) and reconstruction (bottom rows) mode. The model is depicted
in Fig.\ \ref{fig:An-IFT-model} and described by Eqs.\ \eqref{eq:zeta-generation}-\eqref{eq:data-generation}
with the modification that in Eq.\ \eqref{eq:exp-function} the $\exp$-function
is replaced by a sigmoid function to obtain more cloud-like structures.
Run in simulation mode, the model first generates a non-parametric
power spectrum (top right panel) from which a Gaussian realization
of a statistical isotropic and homogeneous field is drawn (top left,
after procession by the sigmoid function). This is then observed tomographically
(top middle), by measurements that integrate over (here randomly chosen)
lines of sight. The data values include Gaussian noise and are displayed
at the locations of their measurement lines. Fed with this synthetic
data set, the model run in inference mode (via geoVI) reconstructs
the larger scales of the signal field (bottom left), the initial power
spectrum (thick orange line in middle right panel; thick blue line
is ground truth), and provides uncertainty information on both quantities
(signal uncertainty is given at bottom middle, the power spectrum
uncertainty is visualized by the set of thin lines at bottom right).
The presented plots are the direct output of the \texttt{getting\_started\_3.py}
script enclosed in the Numerical Information Field Theory (NIFTy)
open source software package NIFTy8, downloadable at \protect\href{https://gitlab.mpcdf.mpg.de/ift/nifty}{https://gitlab.mpcdf.mpg.de/ift/nifty}
\cite{2013A&A...554A..26S,2019AnP...53100290S,2019ascl.soft03008A}
that supports the implementation and inference of IFT models.\label{fig:Output-of-a}}
\end{figure*}

To summarize, the field inference problems of IFT can often be stated
in terms of a standardized, generative model for the signal and the
data. For the illustrative case outlined above, where the probabilistic
model is given by
\begin{eqnarray}
\mathcal{P}(d,\varphi,\psi) & = & \mathcal{P}(d|\varphi)\mathcal{P}(\varphi|\psi)\mathcal{P}(\psi),\label{eq:joint-model}\\
\mathcal{P}(\psi) & = & \mathcal{G}(\psi-\overline{\psi},\Psi),\\
\mathcal{P}(\varphi|\psi) & = & \mathcal{G}(\varphi,\Phi(\psi)),\\
\Phi(\psi) & = & \mathcal{F}^{-1}\widehat{P_{\varphi}}\mathcal{F}^{-1\dagger},\\
P_{\varphi}(k) & = & P_{0}\exp(\psi(\ln(|k|/k_{0})))\text{, and}\\
\mathcal{P}(d|\varphi) & = & \mathcal{P}(d|s(\varphi)),\label{eq:noise-joint-model}
\end{eqnarray}
the corresponding standardized generative model is
\begin{eqnarray}
\zeta & := & (\xi,\eta)\hookleftarrow\mathcal{G}(\zeta,\one),\label{eq:zeta-generation}\\
\psi(\eta) & := & \overline{\psi}+A_{\psi}\eta,\label{eq:psi-generation}\\
P_{\varphi}(k) & := & P_{0}\exp(\psi(\ln(|k|/k_{0})),is\label{eq:Pphi_generation}\\
\varphi(\xi,\psi) & := & \mathcal{F}^{-1}\widehat{P_{\varphi}^{\nicefrac{1}{2}}}\xi,\label{eq:phi-generation}\\
s(\varphi) & := & s_{0}\exp(\varphi),\label{eq:exp-function}\\
n & \hookleftarrow & \mathcal{P}(n|s)\text{, and}\\
d & = & R'(s)+n.\label{eq:data-generation}
\end{eqnarray}
This generative model is illustrated in Fig.\ \ref{fig:An-IFT-model}.
Variants of it are used in a number of real world data applications
\cite{2012A&A...542A..93O,2015A&A...575A.118O,2015A&A...581A..59J,2015MNRAS.449.4162I,2015A&A...581A.126S,2015JCAP...02..041D,2018arXiv180405591K,2019A&A...627A.134A,2020A&A...633A.150H,2020A&A...639A.138L,2020arXiv200205218A,2020arXiv200811435A,2021arXiv210201709H,2021A&A...655A..64M,2014PhRvD..89d3505D,2016JCAP...04..030H,2021JCAP...04..071W}.
Its performance in generative and reconstruction mode is illustrated
for synthetic data in Figs.\ \ref{fig:Output-of-a} and \ref{fig:samples}.

For the noiseless data $d'=R'(s)$ the generative model reads
\begin{eqnarray}
d'(\zeta) & := & R'(s(\varphi(\xi,\psi(\eta))))\nonumber \\
 & = & (R'\circ s\circ\varphi\circ f)(\zeta)\text{, with}\\
f(\zeta) & := & (\xi,\psi(\eta)).
\end{eqnarray}
This way, the full model complexity as given by Eqs.\ \eqref{eq:joint-model}-\eqref{eq:noise-joint-model}
is transferred into an effective response function $d'=R'\circ s\circ\varphi\circ f$.
For this latent variable vector, the prior is simply $\mathcal{P}(\zeta)=\mathcal{G}(\zeta,\one)$,
whereas the likelihood $\mathcal{P}(d|\zeta)=\mathcal{P}(n=d-d'(\zeta)|\zeta)$
has absorbed the full model complexity. This so called reparametrization
trick \cite{2013arXiv1312.6114K} was introduced to IFT by \cite{2018arXiv181204403K}
to simplify numerical variational inference.

At this point it is essential to realize that this generative model
consists of a latent space white noise process $\mathcal{P}(\zeta)=\mathcal{G}(\zeta,\one)$
that generates an input vector $\zeta$ and a sequence of non-local
linear and local non-linear operations that is applied to it. The
Fourier transform $\mathcal{F}^{-1}$ and $A_{\psi}$ are examples
of non-local linear operations within the model. Among the non-linear
operations are the exponential functions and the application of the
$\psi$-dependent amplitude operator $A_{\varphi}(\psi)$ to the latent
space excitations $\xi$, as there the two components of $\zeta=(\xi,\eta)$
are multiplied together. Furthermore, the instrument response $R'(s)$
might also be decomposed into sequences of non-local linear and local
non-linear operations, as physical processes in measurement devices
can often be cast into the propagation of a quantity (an operation
that is linear in the quantity) and the local interactions of the
quantity (an operation non-linear in it), respectively.

\section{Artificial Intelligence\label{sec:Artificial-Intelligence}}

\subsection{Neural Networks}

AI and ML are vast fields. AI aims at building artificial cognitive
systems that perceive their environment, reason about its state and
the systems' best actions, and learn to improve their performance.
ML can be regarded as a sub-field of AI, embracing many different
methods like self-organized maps, Gaussian mixture models, deep neural
networks, and many others. Here, the focus should be on specific neural
networks, GNNs, as those have a close relation to the generative IFT
models introduced before.

GNNs transform a latent space variable $\xi\hookleftarrow\mathcal{G}(\xi,\one)$
into a signal or data realization, $s=s(\xi)$ or $d'=d'(\xi)$. A
neural network is a function $g(\xi)$ that can be decomposed in terms
of $n$ layer processing functions $g_{i}$ with
\begin{equation}
g=g_{n}\circ g_{n-1}\circ\ldots g_{1}.
\end{equation}
Any of the layer processing functions $g_{i}:\xi_{i}\mapsto\xi_{i+1}$
with $\xi_{1}\equiv\xi$ consists typically of a non-local, affine
linear transformation $l_{i}(\xi_{i}):=L_{i}\xi_{i}+b_{i}$ of the
input vector $\xi_{i}$ of layer $i$ followed by a local, point wise
application of non-linear, so-called activation functions $\sigma_{i}:\mathbb{R\rightarrow\mathbb{R}}$.
Thus, the output vector $\xi_{i+1}$ of layer $i$ is 
\begin{eqnarray}
\xi_{i+1} & = & \left(\sigma_{i}\circ l_{i}\right)(\xi_{i}),
\end{eqnarray}
where $\sigma_{i}$ acts component wise. The set $\eta=(L_{i},b_{i})_{i=1}^{n}$
of all coefficients of the $l_{i}$s (the matrix elements of the $L_{i}$
matrices, and the components of the $b_{i}$ vectors) determines the
function the network represents. Putting the input values and network
coefficients into a single vector $\zeta:=(\xi,\eta)$ a GNN can be
regarded as a function of both, latent variables $\xi$ and network
parameters $\eta$, $d'(\zeta)=g(\xi;\eta)$.
\begin{figure*}[t]
\includegraphics[viewport=0bp 864bp 1728bp 1728bp,clip,width=1\textwidth]{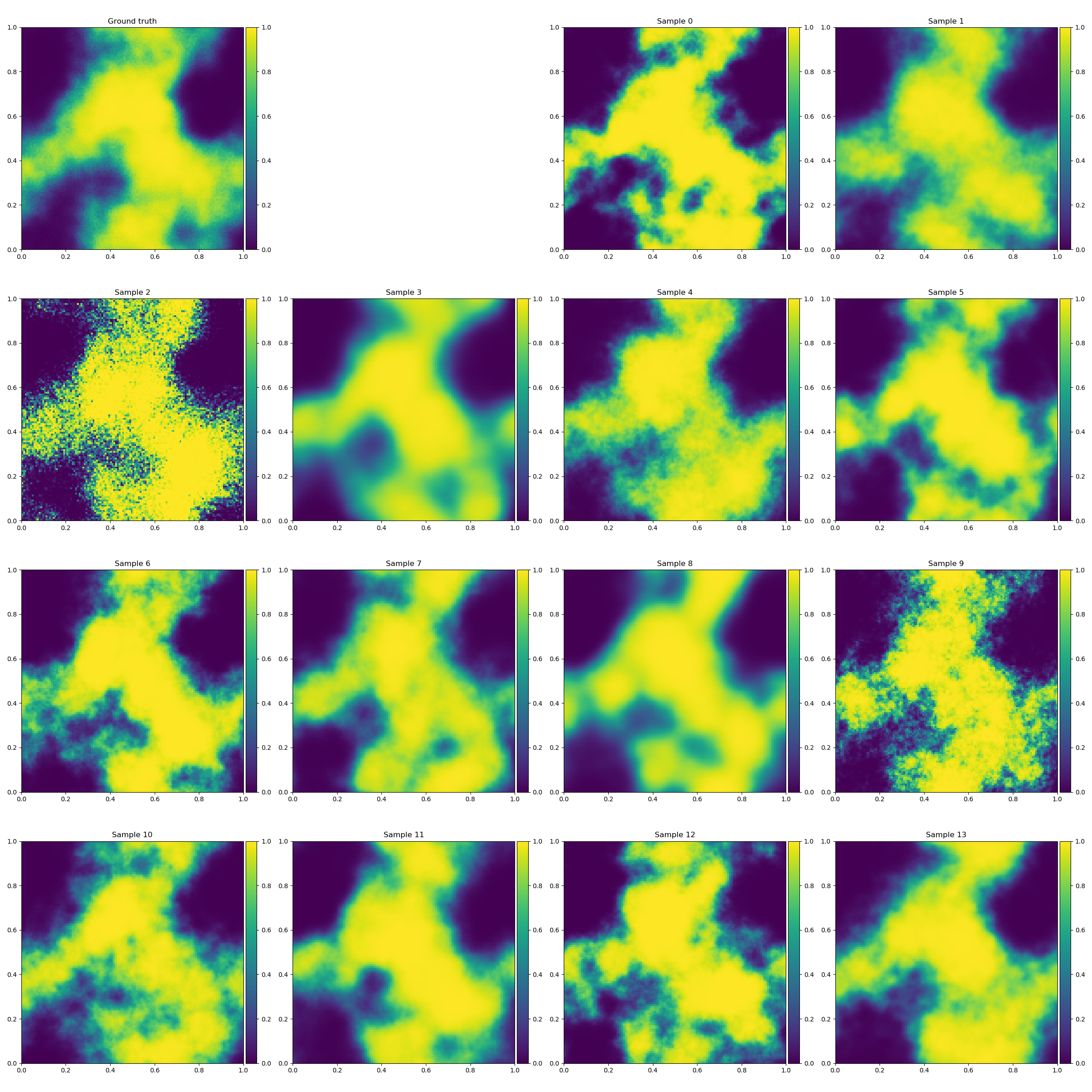}
\caption{Signal ground truth (top left panel) and some signal posterior samples
(other panels) of the field reconstructed in Fig.\ \ref{fig:Output-of-a}.
Note the varying granularity of the field samples due to the remaining
posterior uncertainty of the power spectrum on small spatial scales
as shown in Fig.\ \ref{fig:Output-of-a} at bottom right. \label{fig:samples}}
\end{figure*}

\subsection{Comparison with IFT Models}

From this abstract perspective, a standardized, generative model $d'(\zeta)$
in IFT is structurally a GNN, as both consists of sequences of local
non-linear and non-local linear operations on their input vector $\zeta=(\xi,\eta)$.
The concrete architecture of an IFT model and a typical GNN might
differ significantly, as GNNs often map a lower dimensional latent
space into a higher dimensional data or feature space, whereas the
dimension of the IFT model latent space can be very high, as it contains
a subset of the virtually infinite many degrees of freedom of a field,
see Fig.\ \ref{fig:An-IFT-model}.

Also the way IFT-based models and GNNs are usually used differs a
bit. Both can be used to generate synthetic samples of outputs by
processing random latent space vectors $\xi\hookleftarrow\mathcal{G}(\xi,\one)$.
However, typically an IFT model $d'(\zeta)$ is applied to infer all
latent space variables in $\zeta$ from data $d$. From the latent
variables, the signal of interest can always be recovered via $s(\zeta)$.

For this inference the so-called information Hamiltonian, potential,
or energy 
\begin{eqnarray}
\mathcal{H}(d,\zeta) & = & -\ln\mathcal{P}(d,\zeta)=-\ln\mathcal{P}(d|\zeta)-\ln\mathcal{P}(\zeta)\nonumber \\
 & = & \mathcal{H}(n=d-d'(\zeta)|\zeta)+\frac{1}{2}\zeta^{\dagger}\zeta+\text{const}\label{eq:IFT-loss1}
\end{eqnarray}
is investigated with respect to\ $\zeta$, where $\mathcal{H}(a|b):=-\ln\mathcal{P}(a|b)$.
This quantity is introduced to IFT in analogy to statistical mechanics,
it summarizes the full knowledge on the problem (as it is just a logarithmic
coordinate transformation in the space of probabilities) and has the
nice property, that it allows to speak about information as an additive
quantity, as $\mathcal{H}(a,b)=\mathcal{H}(a|b)+\mathcal{H}(b)$.

Investigating the relevant information Hamiltonian for our IFT problem
$\mathcal{H}(d,\zeta)$ can be done, for example, by minimizing it
to obtain a MAP estimator for $\zeta$ or -- as discussed in the
next section-- via \emph{variational inference} (VI). In case of
a constant, signal independent Gaussian white noise statistics, the
information Hamiltonian becomes 
\begin{eqnarray}
\mathcal{H}(d,\zeta) & = & \frac{|d-d'(\zeta)|^{2}}{2\sigma_{n}^{2}}+\frac{1}{2}\zeta^{\dagger}\zeta+\text{const}.\label{eq:IFT-loss2}
\end{eqnarray}

The training of an usual GNN is done with a training data set $\text{\ensuremath{\widetilde{d}=(d_{i})_{i}}}$
to which a corresponding latent space vector set $\widetilde{\zeta}=(\zeta_{i})_{i}$
and common network parameters $\eta$ need to be found. For this a
loss function of the form
\begin{eqnarray}
\widetilde{\mathcal{H}}(\widetilde{d},\widetilde{\xi},\eta) & = & \sum_{i}\widetilde{\mathcal{H}}(d_{i}|\xi_{i},\eta)+\widetilde{\mathcal{H}}(\widetilde{\xi}|\eta)+\widetilde{\mathcal{H}}(\eta),\label{eq:AI-loss1}\\
\widetilde{\mathcal{H}}(\widetilde{\xi}|\eta) & = & \frac{1}{2}\sum_{i}\xi_{i}^{\dagger}\xi+\text{const}\text{, and}\\
\widetilde{\mathcal{H}}(d_{i}|\xi_{i},\eta) & = & \frac{1}{2}\frac{|d_{i}-d'(\xi_{i},\eta)|^{2}}{2\sigma_{n}^{2}}+\text{const}\label{eq:AI-loss2}
\end{eqnarray}
might be minimized. Here, a typical GNN data loss function $\widetilde{\mathcal{H}}(\widetilde{d}_{i}|\xi_{i},\eta)$
as used for the decoder part of an \emph{autoencoder} (AE) \cite{https://doi.org/10.1002/aic.690370209}
was assumed. In an \emph{generative adversarial network} (GAN) \cite{goodfellow2014generative},
however, this data loss function is given in terms of the output of
a discriminator network. The network parameter prior term $\widetilde{\mathcal{H}}(\eta)$
might be chosen to be uninformative ($\widetilde{\mathcal{H}}(\eta)=\text{const}$)
or informative (e.g. $\widetilde{\mathcal{H}}(\eta)=\frac{1}{2}\eta^{\dagger}\eta$
in case of a Gaussian prior on the parameters).

Anyhow, by comparison of Eqs.\ \ref{eq:AI-loss1}-\ref{eq:AI-loss2}
with Eqs.\ \ref{eq:IFT-loss1} and \ref{eq:IFT-loss2} it should
be apparent that the network loss functions can be structurally similar
to the IFT information Hamiltonian. Both consists of a standardized
quadratic prior-energy and a likelihood-energy and both can have a
probabilistic interpretation in terms of being negative log-probabilities,
e.g. 
\begin{eqnarray}
\mathcal{P}(d,\xi,\eta) & = & e^{-\mathcal{H}(d,\xi,\eta)}\text{ and }\\
\mathcal{P}(\widetilde{d},\widetilde{\xi},\eta) & = & e^{-\widetilde{\mathcal{H}}(\widetilde{d},\widetilde{\xi},\eta)},
\end{eqnarray}
respectively. For this reason, we do not distinguish between an information
Hamiltonian $\mathcal{H}$ and a network loss function $\widetilde{\mathcal{H}}$
by writing $\mathcal{H}$ for both in the following.

The IFT-GNN can operate with solely a single data vector $d$ due
to the domain knowledge coded into their architecture, whereas usual
GNNs require sets of data vectors $\text{\ensuremath{\widetilde{d}=(d_{i})_{i}}}$
to be trained. Recently, more IFT-like architectures for GNNs were
proposed as well, which are also able to process data without training
\cite{2020}.

\section{Variational Inference\label{sec:Variational-Inference}}

\subsection{Basic Idea}

So far, it has been assumed here that MAP estimators are used to determine
network parameters $\zeta$ for both, IFT-based models as well as
traditional GNNs. MAP estimators are known to be prone to over-fitting
the data, as they are not probing the adjacent phase-space volumes
of their solutions. VI methods perform better in that respect, while
still being affordable in terms of computational costs for the high
dimensional settings of IFT-based field inference and traditional
GNN training. They were used in most recent IFT applications \cite{2019A&A...631A..32L,2019A&A...627A.134A,2020A&A...633A.150H,2020A&A...639A.138L,2020arXiv200205218A,2020arXiv200811435A,2021arXiv210201709H,2021A&A...655A..64M,2021JCAP...04..071W}
and are prominently present in the name of \emph{variational autoencoders}
(VAEs) \cite{2013arXiv1312.6114K} that are built on VI.

In VI, the posterior $\mathcal{P}(\zeta|d)$ is approximated by a
simpler probability distribution $\mathcal{Q}(\zeta|d')$, in many
applications by a Gaussian
\begin{equation}
\mathcal{Q}(\zeta|d')=\mathcal{G}(\zeta-\theta,\Theta),
\end{equation}
where $d'=(\theta,\Theta)$. The Gaussian is chosen to minimize the
variational Kullback-Leibler (KL) divergence
\begin{eqnarray}
\text{KL}_{\zeta}(d',d) & := & \mathcal{D}_{\text{KL}}(\mathcal{Q}||\mathcal{P})\nonumber \\
 & = & \int\mathcal{D}\zeta\,\mathcal{Q}(\zeta|d')\ln\frac{\mathcal{Q}(\zeta|d')}{\mathcal{P}(\zeta|d)}\label{eq:VIKL}
\end{eqnarray}
with respect to the parameters of $d'$, $\theta$ and $\Theta$ in
our case.

Ideally, all \emph{degrees of freedom} (DoF) of $\theta$ and $\Theta$
are optimized. In practice, however, this is often not feasible due
to the quadratic scaling of the number of DoF of $\Theta$ with that
of $\theta$. Three approximate schemes for handling the high dimensional
uncertainty covariance will be discussed in the following, leading
to the ADVI, MGVI, and geoVI techniques introduced below, namely
\begin{itemize}
\item mean field theory, in which $\Theta$ is assumed to be diagonal, as
used by ADVI
\item the usage of the Fisher information to approximate $\Theta$ as a
function of $\theta$ and thereby effectively removing the DoF of
$\Theta$ from the optimization problem as used by MGVI
\item a coordinate transformation of the latent space that approximately
standardizes the posterior and therefore sets the covariance to the
identity matrix in the new coordinates, as performed by geoVI.
\end{itemize}
Before these are discussed, a note that applies to all of them is
in order. Optimizing of the VI KL, Eq.\ \eqref{eq:VIKL}, is slightly
sub-optimal from an information theoretical point of view as this
minimizes the amount of information introduced by going from $\mathcal{P}$
to $\mathcal{Q}$. The \emph{expectation propagation} (EP) KL with
reversed arguments $\mathcal{D}_{\text{KL}}(\mathcal{P}||\mathcal{Q})$
would be better, as it minimizes the information loss from approximating
$\mathcal{P}$ with $\mathcal{Q}$ \cite{2016PhRvE..94e3306L}. VI
is known to underestimate the uncertainties, whereas EP conservatively
overestimates them. However, calculating the EP solution for $\theta$
and $\Theta$ would require integrating over the posterior. If this
would be feasible, any posterior quantity of interest could be calculated
as well and there would be no need to approximate $\mathcal{P}(\zeta|d)$
in the first place. Estimating and minimizing the VI KL $\mathcal{D}_{\text{KL}}(\mathcal{Q}||\mathcal{P})$
is less demanding, as the integral over the simpler (Gaussian) distribution
$\mathcal{Q}$ can very often be performed analytically, or by sample
averaging using samples drawn from $\mathcal{Q}$.

\subsection{ADVI and Mean Field Approximation}

In all here discussed VI techniques, the posterior mean $\theta$
and the posterior uncertainty covariance $\Theta$ become parameters
to be determined. The vector $\theta$ has the dimension $N_{\text{dim}}$
of the latent space, whereas the posterior uncertainty covariance
$\Theta$ has $N_{\text{dim}}(N_{\text{dim}}-1)/2=\mathcal{O}(N_{\text{dim}}^{2})$
independent DoF. For small problems, these might be solved for, however,
for large problems with millions of DoF these can not even be stored
in a computer memory. To circumvent this, the \emph{Automatic Differentiation
Variational Inference} (ADVI) algorithm \cite{kucukelbir2017automatic}
often invokes the so called \emph{mean field approximation} (MFA).
This assumes a diagonal covariance $\Theta_{\text{MFA}}=\widehat{\theta'}\equiv\text{diag}(\theta')$,
with $\theta'$ being a latent space vector. Cross-correlations between
parameters can not be represented by this, which is problematic in
particular in combination with the tendency of VI to underestimate
uncertainties.

\subsection{MGVI and Fisher Information Metric}

In order to overcome this limitation of ADVI that limits its usage
in IFT contexts with their large number of DoF, the \emph{Metric Gaussian
Variational Inference} (MGVI) \cite{2019arXiv190111033K} algorithm
approximates the posterior uncertainty of $\zeta$ with the help of
the Fisher information metric 
\begin{eqnarray}
M(\zeta) & := & \left\langle \frac{\partial\mathcal{H}(d|\zeta)}{\partial\zeta}\frac{\partial\mathcal{H}(d|\zeta)}{\partial\zeta}^{\dagger}\right\rangle _{(d|\zeta)}.
\end{eqnarray}
The starting point for obtaining the uncertainty covariance $\Theta$
used in MGVI is the Hessian of the log-posterior 
\begin{eqnarray}
\frac{\partial^{2}\mathcal{H}(\zeta|d)}{\partial\zeta\partial\zeta^{\dagger}} & = & \frac{\partial^{2}\mathcal{H}(d,\zeta)}{\partial\zeta\partial\zeta^{\dagger}}-\underbrace{\frac{\partial^{2}\mathcal{H}(d)}{\partial\zeta\partial\zeta^{\dagger}}}_{=0}\nonumber \\
 & = & \frac{\partial^{2}\mathcal{H}(d,\zeta)}{\partial\zeta\partial\zeta^{\dagger}}
\end{eqnarray}
as a first guess for the approximate posterior precision matrix $\Theta^{-1}$.
Using this evaluated at the minimum $\zeta_{\text{MAP}}$ of the information
Hamiltonian $\mathcal{H}(d,\zeta)$ would correspond to the Laplace
approximation, in which the posterior is replaced by a Gaussian obtained
from doing a saddle point approximation at its maximum.

However, neither is the MAP solution ideal, as discussed above, nor
would this be a good approximation at many locations $\zeta$ that
differ from $\zeta_{\text{MAP}}$. This is because positive definiteness
of the Hessian is not guaranteed there, but it is an essential property
of any correlation and precision matrix. For this reason, $\Theta^{-1}$
can not directly be approximated by this Hessian.

It turns out that the likelihood averaged Hessian is strictly positive
definite, and is therefore a candidate for an approximate posterior
precision matrix for any guessed posterior mean $\theta$. A short
calculation shows that the likelihood averaged Hessian is indeed positive
definite:
\begin{eqnarray}
\Theta^{-1}(\theta) & \approx & \left\langle \frac{\partial^{2}\mathcal{H}(d,\zeta)}{\partial\zeta\partial\zeta^{\dagger}}\right\rangle _{(d|\zeta=\theta)}\nonumber \\
 & = & \left\langle \frac{\partial^{2}\mathcal{H}(\zeta)}{\partial\zeta\partial\zeta^{\dagger}}+\frac{\partial^{2}\mathcal{H}(d|\zeta)}{\partial\zeta\partial\zeta^{\dagger}}\right\rangle _{(d|\zeta=\theta)}\nonumber \\
 & = & \left\langle \one-\frac{\partial^{2}\ln\mathcal{P}(d|\zeta)}{\partial\zeta\partial\zeta^{\dagger}}\right\rangle _{(d|\zeta=\theta)}\nonumber \\
 & = & \one-\left\langle \frac{1}{\mathcal{P}(d|\zeta)\,}\,\frac{\partial^{2}\mathcal{P}(d|\zeta)}{\partial\zeta\partial\zeta^{\dagger}}\right\rangle _{(d|\zeta=\theta)}\nonumber \\
 &  & +\left\langle \frac{1}{\mathcal{P}^{2}(d|\zeta)\,}\frac{\partial\mathcal{P}(d|\zeta)}{\partial\zeta}\frac{\partial\mathcal{P}(d|\zeta)}{\partial\zeta}^{\dagger}\right\rangle _{(d|\zeta=\theta)}\nonumber \\
 & = & \one-\int\text{d}d\,\frac{\cancel{\mathcal{P}(d|\zeta)}}{\cancel{\mathcal{P}(d|\zeta)}}\,\frac{\partial^{2}\mathcal{P}(d|\zeta)}{\partial\zeta\partial\zeta^{\dagger}}\nonumber \\
 &  & +\left\langle \frac{\partial\mathcal{H}(d|\zeta)}{\partial\zeta}\frac{\partial\mathcal{H}(d|\zeta)}{\partial\zeta}^{\dagger}\right\rangle _{(d|\zeta=\theta)}\nonumber \\
 & = & \one-\underbrace{\frac{\partial^{2}}{\partial\zeta\partial\zeta^{\dagger}}\underbrace{\int\text{d}d\mathcal{\,P}(d|\zeta)}_{=1}}_{=0}+M(\theta)\nonumber \\
 & = & \one+M(\theta)>0.\label{eq:ApproxCov}
\end{eqnarray}

The last step follows because the Fisher metric $M(\theta)$ is an
average over outer products ($v\,v^{\dagger}\ge0$) of likelihood
Hamiltonian gradient vectors $v=\partial\mathcal{H}(d|\zeta)/\partial\zeta$
and thereby positive semi-definite. Adding $\one>0$ to the Fisher
metric turns the approximate precision matrix into a positive definite
matrix $\Theta^{-1}(\theta)>0$, of which the inverse $\Theta(\theta)$
exists for all $\theta$, and which is positive definite as well.

\subsection{Exact Uncertainty Covariance}

Being positive definite is of course not the only property an approximation
of the posterior uncertainty covariance has to fulfill. It also has
to approximate well. Fortunately, this seems to be the case in many
situations. The likelihood averaged Laplace approximation actually
becomes the exact posterior uncertainty in case of linear Gaussian
measurement problems as is shown in the following. If it is exact
in such linear situations, it should be a valid approximation in the
vicinity of any linear case.

For linear measurement problems the measurement equation is of the
form $d=R\zeta+n$, the noise statistics $\mathcal{P}(n|\zeta)=\mathcal{G}(n,N)$,
and the standardized prior is $\mathcal{P}(\zeta)=\mathcal{G}(\zeta,\one)$.
The corresponding posterior is known to be a Gaussian
\begin{equation}
\mathcal{P}(\zeta|d)=\mathcal{G}(\zeta-m,D)
\end{equation}
with mean $m$ and covariance $D$ given by the generalized Wiener
filter solution $m=D\,R^{\dagger}N^{-1}d$ and the Wiener covariance
$D=(\one+R^{\dagger}N^{-1}R)^{-1}$, respectively \cite[e.g.][]{2009PhRvD..80j5005E}.
In this case, the Fisher information metric $M=R^{\dagger}N^{-1}R$
is independent of $\zeta$. The approximate posterior uncertainty
covariance as given by Eq.\ \ref{eq:ApproxCov} equals the exact
posterior covariance, $\Theta=\left(\one+M\right)^{-1}=(\one^{-1}+R^{\dagger}N^{-1}R)^{-1}=D$.
Thus indeed, the adopted approximation becomes exact in this situation.
This should motivate why this approximation can hold sensible results
in sufficiently well behaved cases, in particular when a linearization
of the inference problem around a reference solution (e.g. a MAP estimate)
is already a good approximation.

Furthermore, for all signal space directions around this reference
point that are unconstrained by the data, this covariance approximation
returns the prior uncertainty, as it should. Additional discussion
of this approximation can be found in \cite{2019arXiv190111033K},
where also its performance with respect to ADVI is numerically investigated.

The important point about this approximate uncertainty covariance
$\Theta$ is that it is a function of the latent space mean estimate
$\theta,$ \emph{i.e.} $\Theta(\theta)$, and therefore does not need
to be inferred as well. For many likelihoods, the Fisher metric is
available analytically, alleviating the need to store $\Theta$ in
a computer memory as an explicit matrix. It is only necessary that
certain operations can be performed with $\Theta$, like applying
it to a vector or drawing samples from a Gaussian with this covariance.
Relying solely on those memory inexpensive operations the MGVI algorithm
is able to minimize the relevant VI KL, namely $\text{KL}_{\zeta}((\theta,\Theta(\theta)),d)=\mathcal{D}_{\text{KL}}(\mathcal{Q},\mathcal{P})$,
with respect to the approximate posterior mean $\theta$. The result
of MGVI are then the posterior mean $\theta$, the uncertainty covariance
$\Theta(\theta)$, and posterior samples $\left\{ \zeta_{i}\right\} _{i}$
drawn according to this mean and covariance. These samples can then
be propagated into posterior signal samples $s_{i}=s(\zeta_{i})$,
from which any desired posterior signal statistics can be calculated. 

MGVI has enabled field inference for problems, which are too complex
to be solved by MAP, in particular when multiple layers of hyperpriors
were involved \cite[e.g.][]{2020arXiv200205218A,2020A&A...633A.150H}.
A detailed comparison of the performance of MGVI with respect to ADVI
in terms of accuracy and computational speed can be found in \cite{2019arXiv190111033K}.

\subsection{Geometric Variational Inference}

ADVI's and MGVI's weak point, however, can be the Gaussian approximation
of the posterior, which might be strongly non-Gaussian in certain
applications. In order to overcome this, the \emph{geometrical variational
inference} (geoVI) algorithm \cite{2021Entrp..23..853F} was introduced
as an extension of MGVI. geoVI puts another coordinate transformation
on top of the one used by MGVI, so that $\zeta=g_{0}(y)$ -- with
$g_{0}$ to be performed before any of the other IFT-GNN operations
$g_{1},\ldots g_{n}$ -- approximately standardizes the posterior,
$\mathcal{P}(y|d)\approx\mathcal{G}(y,\one)$. Astonishingly, this
transformation can be constructed without the (prohibitive) usage
of any explicit matrix or higher order tensor in the latent space,
thus also allowing to tackle very high dimensional inference problems,
like MGVI. The transformation is basically a normalizing flow (network)
\cite{pmlr-v37-rezende15}, just with the difference to their usual
usage in ML, that the geoVI flow does not need to be trained, but
is derived from the problem statement in form of its information Hamiltonian
in an automated fashion. Specifically, the coordinate transformation
$g_{0}$ is defined to solve the constraining equation
\begin{equation}
\left.\frac{\partial g_{0}}{\partial y}^{\dagger}\Theta(\zeta)\frac{\partial g_{0}}{\partial y}\right|_{\zeta=g_{0}(y)}\approx\one\quad\forall y,
\end{equation}
which fully specifies $g_{0}$ up to an integration constant $\theta$.
This remaining constant is solved for by minimizing the VI KL with
respect to $\theta$ to retrieve the optimal geoVI aproximation.

With geoVI, deeper hierarchical models, which more often exhibit non-Gaussian
posteriors due to a larger number of degenerate parameters in them,
can be approached via VI. The ability of geoVI to provide uncertainty
information is illustrated in Fig.\ \ref{fig:Output-of-a} (bottom
middle and right panels) and in Fig.\ \ref{fig:samples}. Further
details on geoVI and detailed comparisons of ADVI, MGVI, geoVI, and
Hamiltonian Monte Carlo methods can be found in \cite{2021Entrp..23..853F}.

\section{Conclusion and Outlook}

This paper argues that IFT techniques can well be regarded as ML and
AI methods by showing their interrelation with GNNs, normalizing flows,
and VI techniques. This insight is not necessarily new, as this paper
just summarizes a number of recent works \cite{2018arXiv181204403K,2019arXiv190111033K,2021Entrp..23..693K,2021Entrp..23..853F}
that suggested this before.

First, the generative models build and used in IFT are GNNs that can
interpret data without initial training, thanks to the domain knowledge
coded into their architecture \cite{2018arXiv181204403K}. Related
architectures have very recently been proposed as image priors in
the context of neural network architectures as well \cite{2020}.
As IFT models and the newly proposed image priors do not obtain their
intelligence from data driven learning, they are strictly not ML techniques,
but might be characterized as (expert) knowledge driven AI systems.
From a technical point of view, however, such a distinction could
be seen as splitting hairs.

Second, the VI algorithms used in IFT and AI to approximately infer
quantities are a natural interface between these areas. Here, the
related ADVI \cite{kucukelbir2017automatic}, MGVI \cite{2019arXiv190111033K},
and geoVI \cite{2021Entrp..23..853F} algorithms were briefly discussed,
which can be used in classical ML and AI as well as in IFT applications.

And third, the common probabilistic formulation of IFT models and
GNNs, as well as the common VI infrastructure of the two areas allows
for combining pre-trained GNNs and other networks with IFT-style model
components. In that respect, the possibility to perform Bayesian reasoning
with trained neural networks as described in \cite{2021Entrp..23..693K}
might give an outlook on the potential to combine IFT with other ML
and AI methods.

To summarize, IFT \cite{2009PhRvD..80j5005E,2019AnP...53100127E}
addresses perception \cite{2012A&A...542A..93O,2015A&A...575A.118O,2015A&A...581A..59J,2015MNRAS.449.4162I,2015A&A...581A.126S,2015JCAP...02..041D,2018arXiv180405591K,2019A&A...627A.134A,2020A&A...633A.150H,2020A&A...639A.138L,2020arXiv200205218A,2020arXiv200811435A,2021arXiv210201709H,2021A&A...655A..64M,2014PhRvD..89d3505D,2016JCAP...04..030H,2021JCAP...04..071W},
reasoning \cite{2012PhRvE..85b1134S,2013PhRvE..87a3308E,2018PhRvE..97c3314L,2019Entrp..22...46K,2021AnP...53300486F,2021Entrp..23..693K},
and adaptive inference \cite{2019arXiv190111033K,2021Entrp..23..853F}
tasks. All these are central to the aims of AI and ML to build intelligent
systems including such for perception, cognition, and learning.

\label{sec:Conclusion-and-Outlook}
\begin{acknowledgement*}
I am grateful to many colleagues and students that helped me to understand
the relation of IFT and AI. The line of thoughts presented here benefited
particularly from discussions with Philipp Arras, Philipp Frank, Jakob
Knollmüller, and Reimar Leike. I thank Philipp Arras, Vincent Eberle,
Gordian Edenhofer, Johannes-Harth-Kitzerow, Philipp Frank, Jakob Roth,
and three constructive anonymous reviewers for detailed comments on
the manuscript.
\end{acknowledgement*}
\printbibliography

\end{document}